\documentclass{article}

\usepackage[preprint, nonatbib]{preprint}

\usepackage[utf8]{inputenc} 
\usepackage[T1]{fontenc}    
\usepackage{hyperref}       
\usepackage{url}            
\usepackage{booktabs}       
\usepackage{amsfonts}       
\usepackage{nicefrac}       
\usepackage{microtype}      
\usepackage{xcolor}         
\usepackage{graphicx}
\usepackage{amsmath}
\date{}

\title{Aiding Medical Diagnosis through Image Synthesis and Classification}

\author{%
  Kanishk~Choudhary \\
  Independent Researcher\\
  Fremont, CA 94539 \\
  \texttt{kanishkc2020@gmail.com} \\
}

\begin{document}

\maketitle

\begin{abstract}
    Medical professionals, especially those in training, often depend on visual reference materials to support an accurate diagnosis and develop pattern recognition skills. However, existing resources may lack the diversity and accessibility needed for broad and effective clinical learning. This paper presents a system designed to generate realistic medical images from textual descriptions and validate their accuracy through a classification model. A pretrained stable diffusion model was fine-tuned using Low-Rank Adaptation (LoRA) on the PathMNIST dataset, consisting of nine colorectal histopathology tissue types. The generative model was trained multiple times using different training parameter configurations, guided by domain-specific prompts to capture meaningful features. To ensure quality control, a ResNet-18 classification model was trained on the same dataset, achieving 99.76\% accuracy in detecting the correct label of a colorectal histopathological medical image. Generated images were then filtered using the trained classifier and an iterative process, where inaccurate outputs were discarded and regenerated until they were correctly classified. The highest performing version of the generative model from experimentation achieved an F1 score of 0.6727, with precision and recall scores of 0.6817 and 0.7111, respectively. Some types of tissue, such as adipose tissue and lymphocytes, reached perfect classification scores, while others proved more challenging due to structural complexity. The self-validating approach created demonstrates a reliable method for synthesizing domain-specific medical images because of high accuracy in both the generation and classification portions of the system, with potential applications in both diagnostic support and clinical education. Future work includes improving prompt-specific accuracy and extending the system to other areas of medical imaging.
\end{abstract}

\section{Introduction}
Medical imaging is a foundational tool for clinical diagnosis and education. Yet, access to a diverse range of high-quality medical images is often limited. Trainees may rely on only a few reference cases and rarely encounter certain conditions, hindering their learning. This creates a need for systems that can generate accurate, representative medical images on demand to augment training datasets and support diagnostic reasoning [1, 2]. I focus on generating realistic images for human training rather than solely developing classification models because human medical practitioners remain central to clinical decision-making due to reasons such as interpretability, patient trust, regulatory requirements, and liability concerns. Generative models have shown promise in producing synthetic medical images, but the precision required in medicine means that any inaccuracy can be misleading or harmful. My goal is to produce synthetic images with both the realism and the semantic accuracy necessary for clinical use.

Generative Adversarial Networks (GANs) have been widely used for medical image synthesis and augmentation [1]. Goodfellow et al.’s GAN framework [2] trains a generator against a discriminator to create realistic images, and numerous studies have applied GANs to expand medical training data [1, 3]. For example, Frid-Adar et al. improved liver lesion classification by adding GAN-generated images to a training set, boosting a classifier’s performance [10]. However, GANs often require large specialized datasets and careful tuning [1], and they commonly suffer from instability and mode collapse on limited data [1].

Diffusion models have emerged as a compelling alternative, achieving high-resolution, semantically rich outputs that rival or surpass GANs [4]. Latent diffusion models (e.g., Stable Diffusion) generate images via iterative denoising in a latent space [5], demonstrating state-of-the-art results on natural image synthesis [4]. Yet, a model like Stable Diffusion, trained on broad datasets (e.g., LAION-5B), performs poorly on specialized medical imagery without fine-tuning [6]. If prompted with medical terms out-of-the-box, it may produce anatomically incorrect results (Figure 1) [6]. This gap has spurred efforts to adapt diffusion models to medical domains [6].

This work introduces a self-validating generative framework to address both realism and accuracy in medical image synthesis. I fine-tuned Stable Diffusion v1.5 on a histopathology dataset (PathMNIST [9]) using LoRA [7] to specialize it for colorectal tissue images. I then trained a ResNet-18 classifier [8] on the same dataset to serve as an automated validator. During generation, each output image is evaluated by the classifier; if the predicted class does not match the input prompt, the image is discarded and the model tries again. This iterative loop continues until a generated image is classified correctly, ensuring that only images consistent with the target class are retained. My two-model pipeline (generator + validator) thus enforces semantic accuracy in real time. The result is a set of synthetic images that are not only visually realistic but also labeled correctly, meeting the high standards required for clinical and educational use.

\begin{figure}[ht]
\centering
\includegraphics[width=0.8\linewidth]{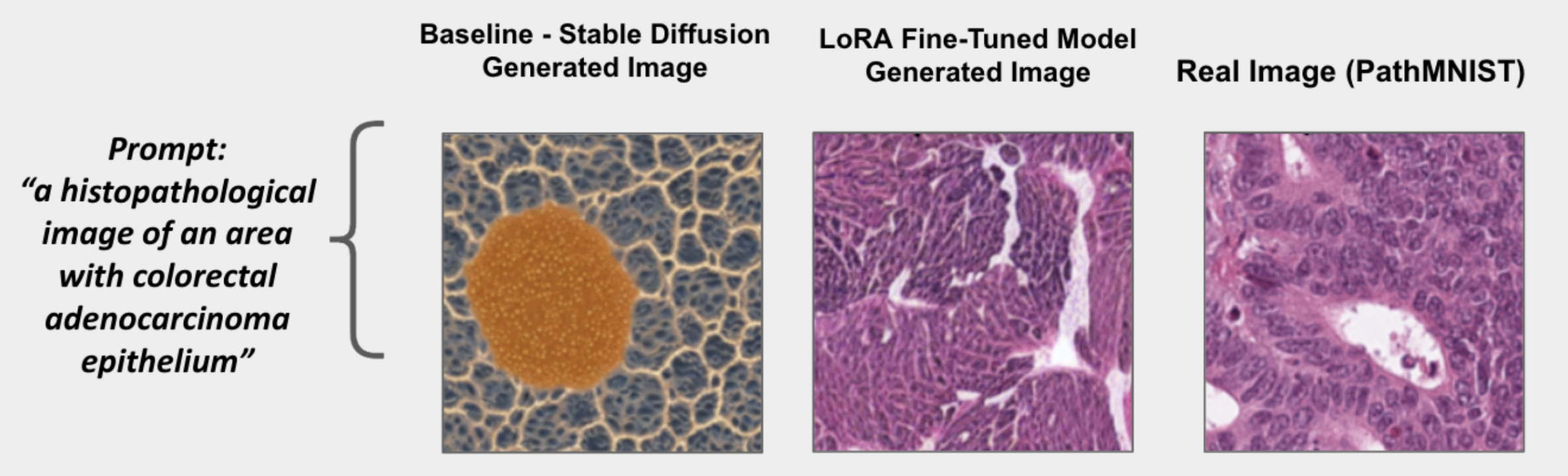}
\caption{Example of Generated vs. Real Images: Comparison between a real histopathology image (right), a synthetic image generated by my system (middle), and an inaccurate image generated by a baseline Stable Diffusion model (left). These images all aim to represent the same tissue type.}
\label{fig:generatedvsreal}
\end{figure}

\section{Background}
Combining image generation with automated validation in the medical domain is relatively new. Most prior studies focus on generating realistic images and evaluating them indirectly (e.g., by seeing if synthetic data improves a model’s performance) rather than checking each image’s accuracy. GAN-based methods have dominated early work in medical image synthesis [1]. Frid-Adar et al. pioneered using GANs to generate synthetic liver lesions, which improved a classifier when added to the training set. This demonstrated synthetic data’s potential to bolster limited datasets, but generation quality was evaluated only via downstream task performance rather than by verifying each image.

More recently, Xue et al. introduced HistoGAN [11], a conditional GAN for histopathology that includes a selective augmentation step. HistoGAN generates images conditioned on class labels and then uses a classifier to filter out low-confidence outputs, adding only the most realistic samples to the training set. This provides a form of post-hoc quality control and leads to improved histology image classification. However, HistoGAN’s validation occurs after generation—to decide which images to keep for augmentation—rather than intervening during the image creation process.

Other approaches evaluate synthetic images via classifier metrics or expert review. For instance, some diffusion-based pipelines gauge realism with pathologist surveys and measure how well models trained on synthetic images perform [3]. These underline the importance of quality in medical image synthesis, yet they stop short of enforcing correctness during generation.

My approach tightly integrates generation and validation. Unlike prior GAN or diffusion studies that optimize for visual realism or rely on indirect validation (e.g., improved downstream task performance), my system explicitly checks each output against the target label as it is produced. Indirect validation provides only aggregate feedback and cannot pinpoint specific failed images. In contrast, my self-validation mechanism forces the generator to immediately regenerate or discard any output the classifier deems incorrect. Conceptually, this is similar to classifier-guided generation, but rather than using the classifier’s gradients to steer the generation process (as in some diffusion models [4]), I use its predictions to iteratively accept or reject outputs until the criteria are met in real time.

I fine-tuned a state-of-the-art diffusion model on a specialized dataset and paired it with a domain-trained validator, creating an automated feedback loop for semantic accuracy in synthetic images. To my knowledge, this is one of the first systems to actively enforce ground-truth consistency (each generated image’s class matches its prompt) during image generation, rather than treating validation as an afterthought. I show that this framework produces synthetic images that are not only visually plausible but also credibly representative of their intended class. Such images can serve as a reliable resource for data augmentation, medical training, and even diagnostic support in scenarios where real data are scarce.

\section{Methods}

\subsection{Dataset}
I conducted my experiments on the PathMNIST dataset (part of MedMNIST v2 [9]), which consists of 107,180 hematoxylin-eosin stained pathology image patches categorized into nine tissue classes. These classes correspond to distinct colorectal tissue types (e.g., adipose, background, debris, lymphocytes, mucus, smooth muscle, normal colon mucosa, cancer-associated stroma, and adenocarcinoma epithelium) derived from the NCT-CRC-HE-100K collection [12]. Each image is a small region of a histology slide, resized to 224×224 pixels to standardize the input dimensions. I used an 80/20 split of the dataset, allocating 80\% of the images for training and 20\% for testing. Within the training set, a portion was further set aside for internal validation during model development. This split was stratified to preserve class balance – each tissue type is represented by on the order of ~10,000 training examples – thereby avoiding severe class imbalance. For the classification model, I applied data augmentation techniques (random rotations, zooms, contrast adjustments, etc.) to the training images. These augmentations improve the classifier’s robustness to image variability without altering the underlying tissue label. Importantly, I did not apply augmentations when fine-tuning the generative model, since I wanted the diffusion model to learn each class’s true visual characteristics without augmented distortions (ensuring that the text prompt “lymphocytes” always maps to authentically appearing lymphocyte patches, for example).

\subsection{System Architecture}
My system comprises two main modules – a generative module and a validation module – integrated in an iterative feedback loop. The generative component is built on Stable Diffusion v1.5 [3], a latent diffusion model capable of producing high-resolution images from text prompts. I fine-tuned this model on the PathMNIST training set using the LoRA technique [4], which introduces a small number of trainable low-rank adaptation parameters into the network’s attention layers to efficiently specialize the model to my domain. The validation component is a ResNet-18 classifier trained on the PathMNIST data to recognize the nine tissue classes with near-perfect accuracy. If the classifier’s predicted label for a generated image does not match the conditioning prompt, the image is discarded and the diffusion model regenerates a new sample. This integrated design ensures that the outputs are not only visually realistic but also semantically correct for the target class.

\begin{figure}[ht]
\centering
\includegraphics[width=0.8\linewidth]{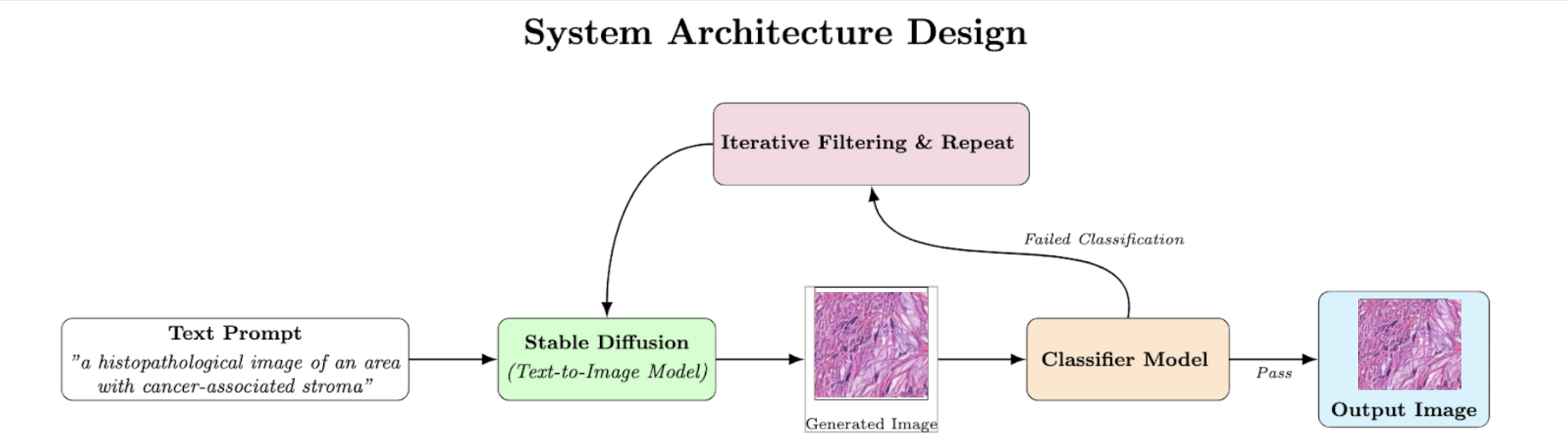}
\caption{System diagram of the self-validating medical image synthesis pipeline: 
Prompts guide image generation; generated outputs are validated by a classifier; failed generations are discarded and regenerated until the label is correct.}
\label{fig:systemdesign}
\end{figure}

\subsection{Experimentation Setup}
I implemented the system in Python (using PyTorch) and conducted all training and evaluation in a managed Jupyter environment (Google Colab). I leveraged two GPU types during development: an NVIDIA A100 (40 GB) for computationally intensive tasks (e.g., diffusion model fine-tuning at higher resolution) and an NVIDIA T4 (16 GB) for lighter workloads (e.g., prototyping, classifier training, and image generation). The A100’s larger memory enabled training with bigger batch sizes, while the T4 was sufficient for inference and for training the smaller ResNet-18 model.

To ensure reproducibility, I employed version control and thorough experiment logging. All code, model configurations, and training logs were tracked in documents and a GitHub repository. I used systematic naming/versioning (V1–V10 for different diffusion model variants) to organize my experiments, and saved model checkpoints at regular intervals (every few hundred steps) to enable rollback or analysis of intermediate results. I also developed automated evaluation scripts to compute key metrics (precision, recall, F1-score, and confusion matrices) on batches of generated images using the classifier, as well as a utility to generate a fixed set of prompts covering all tissue classes for consistent testing. These tools and practices ensured that my entire workflow is reproducible and that any researcher can obtain the same results by following the provided code and settings.

\subsection{Experimentation Details}
\paragraph{Training Strategy and Model Versioning}
I iteratively fine-tuned the diffusion model through multiple versions (V1–V10), systematically adjusting key hyperparameters to improve performance. For example, I experimented with different training durations (around 1,000 to 3,000 steps), batch sizes (8, 16, 32), and learning rates (on the order of 1e-5 to 5e-5, with gradual warm-up) to balance underfitting and overfitting. I also applied best practices such as a small noise offset during diffusion training and a cosine learning-rate decay schedule, which improved training stability and image quality. Each intermediate model was evaluated on a validation set of generated images using the classifier-based metrics, guiding the selection of subsequent variants. I observed a steady upward trend in accuracy over these versions: for instance, an early Version 6 achieved a macro F1 of around 0.43, which improved to around 0.54 by Version 8. The best model (Version 9 at 1131 steps) attained the highest performance (precision 0.68, recall 0.71, F1 0.67) and was chosen as the final deployment (see Figure 3).

\begin{figure}[ht]
\centering
\includegraphics[width=0.8\linewidth]{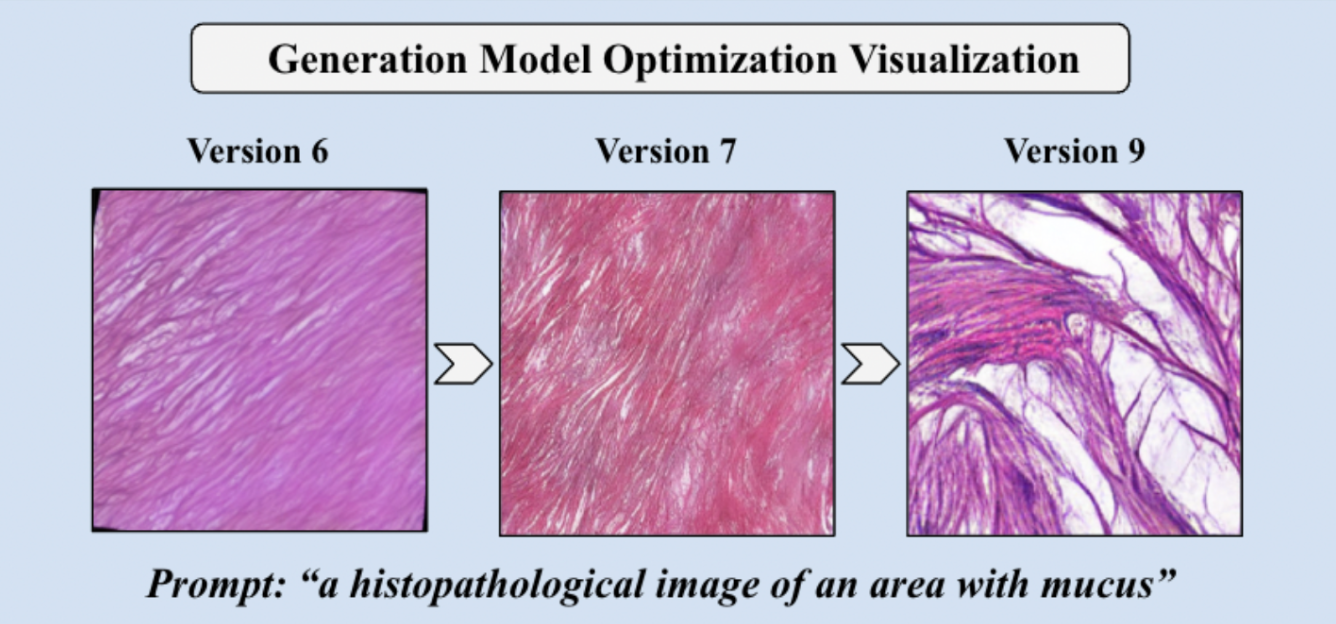}
\caption{Shows progression of accuracy of the generation model as more versions were trained and improved on: Version 9’s results for a prompt relating to mucus were much more accurate than Versions 6 and 7.}
\label{fig:optimization}
\end{figure}

\paragraph{Evaluation Procedure}
To quantitatively assess the quality of the generated images, I evaluated each model version using classifier-based metrics. For each experiment, the diffusion model was prompted to generate a fixed number of images per tissue class (e.g., 10 images for each of the 9 classes), and I used the ResNet-18 validator to predict a label for each generated image. I then compared these predictions to the known prompt labels to compute precision, recall, and F1-score for that generative model. In my context, precision is the fraction of generated images correctly classified as the intended class, and recall is the fraction of intended images that were successfully generated on the first attempt. The F1-score (the harmonic mean of precision and recall) provides an overall measure of generation accuracy. I report these metrics as macro-averages across all classes. This evaluation procedure, combined with the regeneration loop (which retries generation until the output is classified correctly), provides a robust assessment of the system’s performance.

\section{Results}
\subsection{Comparison To Baseline}
I first compared my fine-tuned diffusion model to the original pretrained model (without domain adaptation). The baseline Stable Diffusion model (no fine-tuning) produced mostly unrealistic pathology images, where its overall F1-score was only 0.122, indicating it rarely generated a correctly classifiable image. Visually, the baseline outputs often contained nonsensical textures or mixed tissue features (Figure 4, left). After fine-tuning with LoRA and incorporating the self-validation loop, the quality improved dramatically. Many synthetic images from my model are indistinguishable from real histology (Figure 4, middle vs. right), and the classifier confirms their correctness.

\begin{figure}[ht]
\centering
\includegraphics[width=0.8\linewidth]{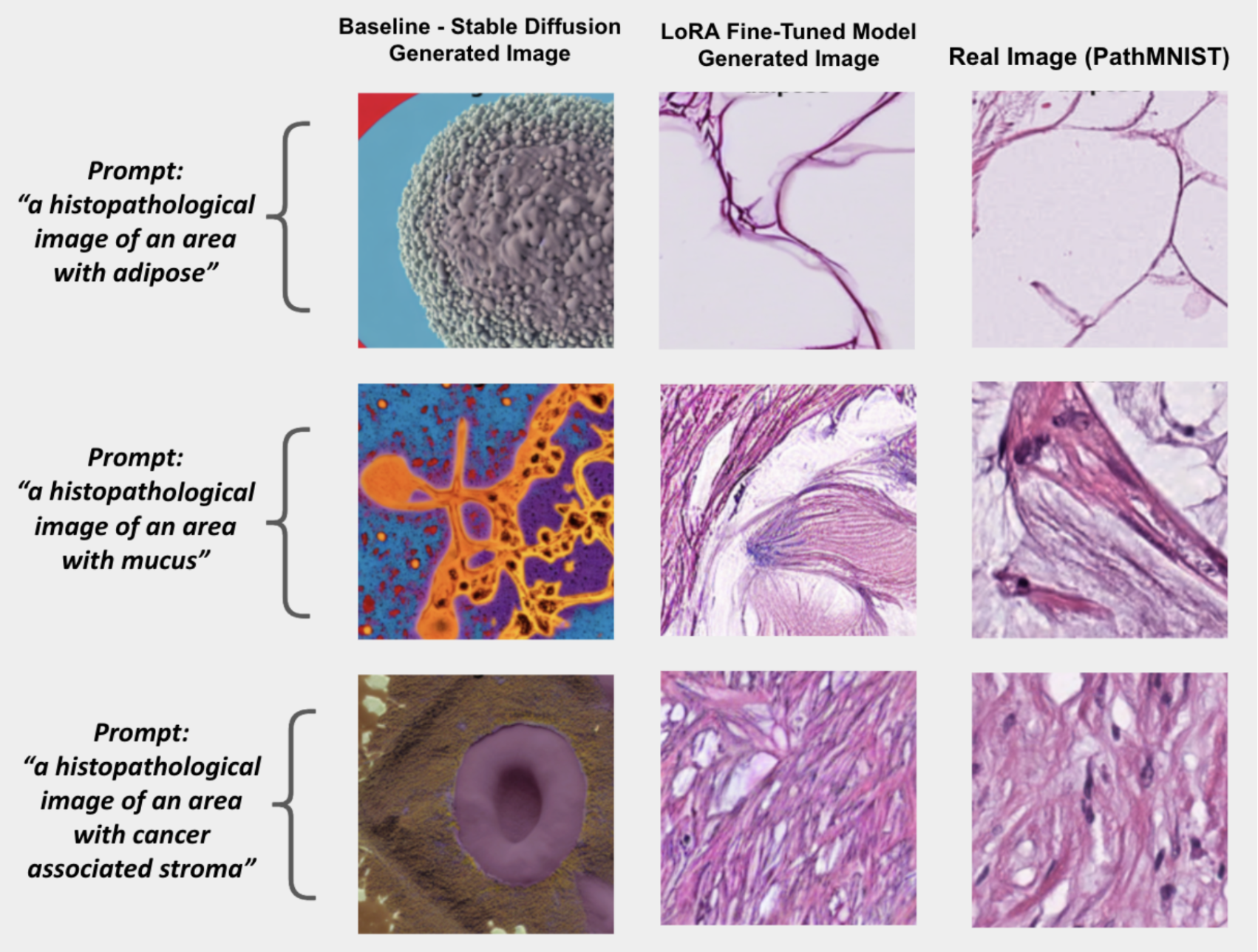}
\caption{Real vs. Generated Histopathology Images: Example outputs for the same prompts using the baseline Stable Diffusion model (left), my fine-tuned LoRA model (middle), and real PathMNIST images (right). The baseline model’s images are anatomically implausible, whereas the fine-tuned model produces realistic glandular structures closely matching the ground truth.}
\label{fig:extra}
\end{figure}

\subsection{Quantitative Performance of Generated Images}
I evaluated my generative model using the classifier’s performance on a set of generated images (ten images per class). Overall, the fine-tuned model achieved an average F1-score of 0.6727 across the nine classes, with precision 0.6817 and recall 0.7111. This is a substantial improvement over the unadapted model’s F1 ~0.12. It also outperformed an early fine-tuning attempt without some of my training edits (which reached ~0.54 F1). In practical terms, the system generates a correct image majority of the time on the first attempt; the self-validation loop handles the rest by regenerating until success.

I also examined performance per class. Some tissue categories are synthesized with very high fidelity: for example, adipose tissue, lymphocytes, and debris achieved F1-scores near 0.9–1.0. This means the generator produces those tissues almost flawlessly according to the classifier (indeed, adipose outputs were 100\% correctly classified). On the other hand, a few classes remained challenging. Mucus, adenocarcinoma epithelium, and normal colon mucosa had the lowest F1-scores, indicating confusion between those classes. Indeed, the confusion matrix of the classifier’s predictions on the synthetic images (Figure 6) shows that some normal colon mucosa samples were misclassified as cancer-associated stroma. These errors highlight where the generative model struggles with fine-grained distinctions.

\begin{figure}[ht]
\centering
\includegraphics[width=0.8\linewidth]{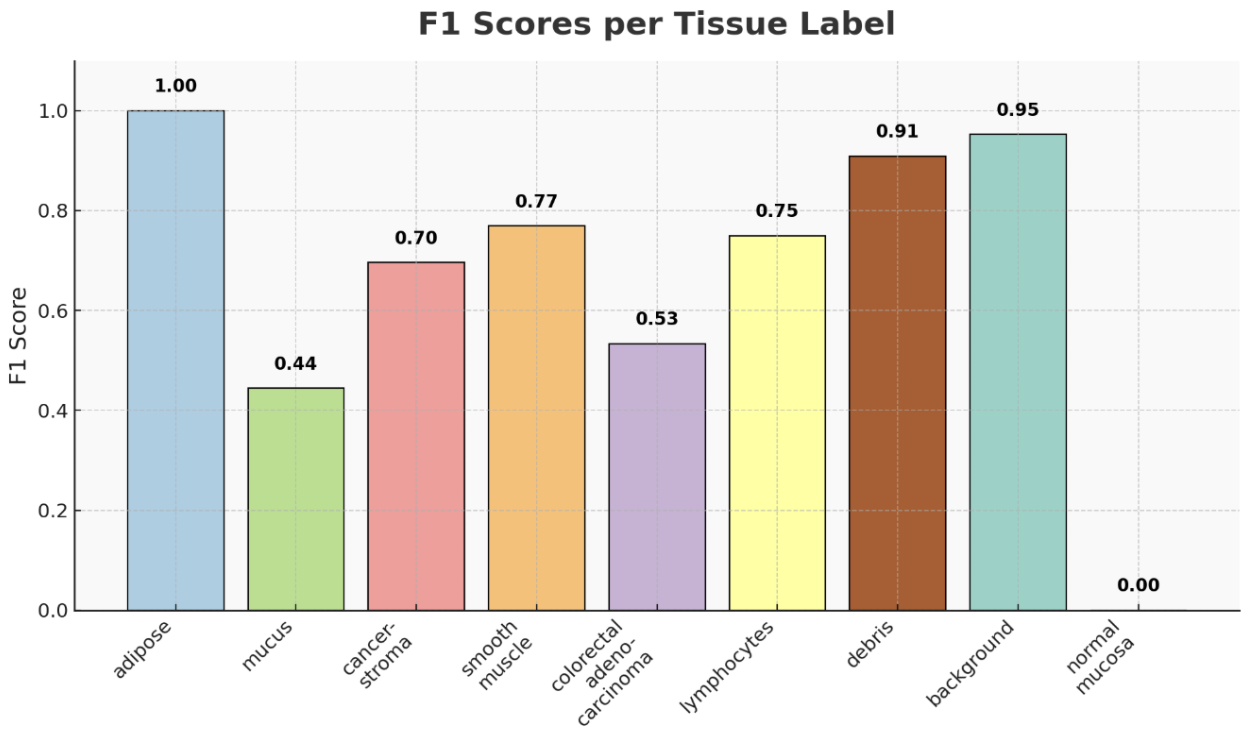}
\caption{Per-Class F1 Scores for Best Model: Simpler tissue classes achieve very high F1 (near 1.0), while more complex classes (mucus, normal colon mucosa) have lower scores.}
\label{fig:bargraph}
\end{figure}

To further analyze the performance, I constructed a confusion matrix of the classifier’s decisions on the best model’s outputs.

\begin{figure}[ht]
\centering
\includegraphics[width=0.8\linewidth]{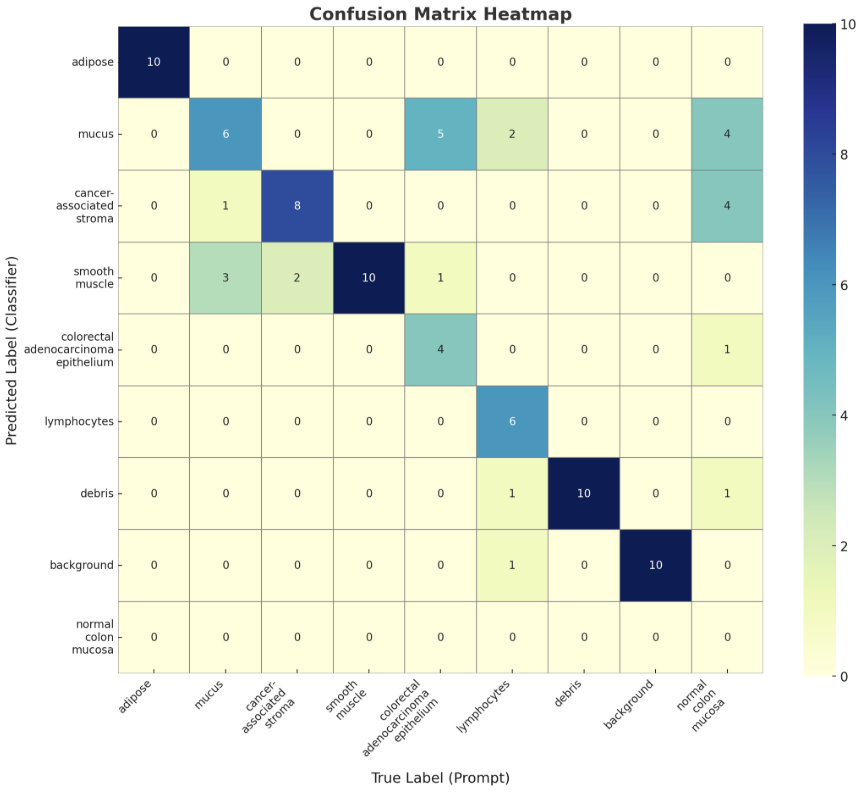}
\caption{Confusion Matrix for Generated Image Classification (Version 9, Step 1131). Confusion matrix of the classifier’s results on synthetic images (best model). Each cell shows the count of generated images of a true class (columns) classified as each predicted class (rows). Most images fall along the diagonal (correct), with off-diagonal errors highlighting confusion.}
\label{fig:confusionmatrix}
\end{figure}

\subsection{Overall Effectiveness}
Taken together, my results demonstrate that the self-validating generation approach can reliably produce realistic and accurate synthetic histopathology images across a range of tissue types. When the model generates a clear example (which it does on the first try in most cases), the classifier simply verifies it. If the model’s output is ambiguous or incorrect, the self-validation loop serves as a safety net, prompting the model to try again until the output meets the diagnostic criteria. This process ensures that the final collection of images is high-quality and label-consistent. In practice, the system could be used to build a library of synthetic images with the confidence that each image depicts the intended tissue correctly.

My approach also provides insight into its failure modes. By examining misclassified outputs via the confusion matrix, I can identify which classes need improvement. For instance, noticing that “mucus” images are often mistaken for adenocarcinoma suggests the need for more training examples of mucus or more descriptive prompts to better distinguish those classes. This kind of built-in self-auditing is an advantage: the system not only generates data but also flags where it is less reliable, guiding future model refinements. Overall, the self-validating framework proves to be a viable approach for producing synthetic medical images that maintain both realism and semantic accuracy, making them useful for data augmentation and education.

\section{Conclusions}
I presented a novel generative framework for the medical domain that emphasizes reliability and diagnostic accuracy. The system effectively bridges the gap between general-purpose image synthesis and the strict precision needed in clinical contexts. General diffusion models operate over a broad, unbounded visual space, which introduces a high risk of medically irrelevant or incorrect outputs. Even minor errors in a synthetic medical image can have significant consequences. My approach narrows this space to the medically valid subset by fine-tuning on domain-specific data and enforcing correctness through an integrated validator.

My results indicate strong performance in generating and validating images for simpler tissue classes (like adipose and debris), while more complex textures (such as mucus and certain tumor patterns) remain challenging. This highlights areas for improvement, such as refining prompt design, obtaining more training examples for underrepresented classes, or exploring alternative model architectures to better capture subtle differences. Importantly, I observed cases where an image looked plausible to the human eye but was caught by the classifier as incorrect, underscoring the value of objective, model-based quality control.

Although my study focused on histopathology, the modular design of the framework allows it to be adapted to other medical imaging domains with minimal changes. This generalizability means the approach could be scaled to various high-need areas, providing a consistent method to generate abundant, reliable synthetic data.

\section{Future Work}
While my framework performs well overall, there is room for further improvement. I plan to address the underperforming categories (such as mucus and normal colon mucosa) by obtaining additional training data or employing targeted fine-tuning strategies for those classes. Better prompt engineering or class-conditional guidance might help the model differentiate challenging tissue types. Another direction is to incorporate more advanced validation criteria—such as multiple expert-trained classifiers or morphological consistency checks—to catch subtler errors. Making image generation more robust to various prompt structures is another pathway to explore. I also aim to extend the system to other medical imaging tasks. For instance, adapting the pipeline to radiology (X-rays, MRIs) or dermatology could provide synthetic data in areas where patient privacy or data scarcity is a concern. Such extensions would validate the framework’s versatility. Ultimately, scaling up this approach and integrating it into clinical AI pipelines could amplify its impact: imagine an on-demand generator of rare disease examples for training clinicians, or a reliable synthetic data supplement for machine learning models in domains where real annotated data are hard to come by. By continually refining the generation and validation components, we can move closer to that vision.

\end{document}